\documentclass[11pt,a4paper]{article}
\usepackage[hyperref]{acl2019}
\usepackage{times}
\usepackage{latexsym}
\usepackage[T2A,T1]{fontenc}
\usepackage[utf8]{inputenc}
\usepackage[russian,english]{babel}

\usepackage{adjustbox}
\usepackage{amsmath, amssymb}
\usepackage{booktabs}
\usepackage{CJKutf8}
\usepackage{dirtytalk}
\usepackage{enumitem}
\usepackage{flushend}
\usepackage{mathptmx}
\usepackage{microtype}
\usepackage{supertabular}
\usepackage[T1]{tipa}
\usepackage{url}

\newcommand{\term}[1]{\textbf{#1}}
\newcommand{\lang}[1]{\texttt{#1}}
\newcommand{\foreign}[1]{\textit{#1}}

\def\Snospace~{\S{}}

\aclfinalcopy

\defcitealias{berlin1969basic}{B\&K}

\title{Modeling Color Terminology Across Thousands of Languages}

\author{
    Arya D. McCarthy\textnormal{,} 
    Winston Wu\textnormal{,} 
    Aaron Mueller\textnormal{,}\\ 
    \textbf{
        Bill Watson\textnormal{, and} 
        David Yarowsky
    }\\
    Department of Computer Science\\
    Johns Hopkins University, Baltimore, MD USA\\
    \texttt{\{arya,wswu,amueller,billwatson,yarowsky\}@jhu.edu} 
}

\date{}

\begin{document}
\maketitle
\begin{abstract}
There is an extensive history of scholarship into what constitutes a ``basic'' color term, as well as a broadly attested acquisition sequence of basic color terms across many languages, as articulated in the seminal work of \citet{berlin1969basic}. This paper employs a set of diverse measures on massively cross-linguistic data to operationalize and critique the \citeauthor{berlin1969basic} color term hypotheses.
Collectively, the 14 empirically-grounded computational linguistic metrics we design---as well as their aggregation---correlate strongly with both the \citeauthor{berlin1969basic} basic/secondary color term partition (\(\gamma=0.96\)) and their hypothesized universal acquisition sequence. The measures and result provide further empirical evidence from computational linguistics in support of their claims, as well as additional nuance: they suggest treating the partition as a spectrum instead of a dichotomy.
\end{abstract}

\section{Introduction}  

How many colors are in the rainbow?
An infinite number, but each language divides up perceptual space into a \emph{finite} number of categories by giving names to colors.
The seminal work on color categories, by \citet[hereafter \citetalias{berlin1969basic}]{berlin1969basic}, characterizes a \emph{universal evolutionary sequence} for languages' core colors (their \term{basic color terms}) and their corresponding categories, at each stage refining the partition of color space. 

A handful of criteria define basic color terms, including abstractness, monomorphemicity, and not being subsumed by a broader basic term. (See \autoref{sec:bcts} for the complete list.) These criteria are accused of biasing analyses of color systems---especially in non-Western societies \citep{wierzbicka2006semantics}. To mitigate this bias, a pan-lingual approach to analyzing color systems may reveal general (\say{universal}) trends more reliably 
than smaller datasets. While data are hard to find in the long tail of languages, we still aim to consider more than ever before---2491 languages and dialects.\footnote{ To this end, we present a 
large cross-lingual, type-level database of translations of basic and secondary color terms across 2491 languages (\autoref{sec:data}).} We leverage natural language processing tools to operationalize longstanding literature on language universals. 

\begin{table}
    \centering
    \begin{tabular}{l l l}
    \toprule
        Language & Color Word & Literal Gloss \\
    \midrule
        Welsh & brown & brown \\
        Italian & marrone & chestnut \\
        Persian & \includegraphics{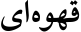} & coffee + of \\
        Cantonese & \begin{CJK}{UTF8}{gbsn}啡色\end{CJK} & coffee + color \\
                
    \bottomrule
    \end{tabular}
    \caption{Examples of terms representing \emph{brown}, arising from four processes: 
    borrowing (Welsh; from English), null affixing (Italian), derivational affixing (Persian), and compounding (Cantonese).}
    \label{tab:intro}
\end{table}

%
%
We provide a three-pronged investigation of the classic criteria for basic color terms, examining the degree to which color words are abstract (\autoref{sec:abstractness}), monomorphemic/monolexemic (\autoref{sec:morphology}), and salient (\autoref{sec:salience}). 
Our operationalization of these \citepalias{berlin1969basic} criteria shows that individual features do not reflect the basic/non-basic divide. Nor is this divide binary, as \citetalias{berlin1969basic} suggest: 
We show that abstractness, monomorphemicity, and even salience do not cleanly divide colors. 

Nonetheless, by treating basicness as a spectrum and aggregating these features (like human-judged concreteness, frequency of compounding, and word length) into basicness scores (\autoref{sec:aggregation}), we can largely distinguish between basic and non-basic colors (validating our measures), and \term{our scores recreate the historical sequence of color acquisition in language}. The sequence is in no way directly encoded in the criteria for basic color terms; as such, recreating it is a separate and novel empirical discovery.

\section{Color Terminology} \label{sec:bcts}

Not all languages have the same number of color words; for instance, a single Korean color word (\foreign{pureu-n}) applies to both grass and sky---an unusual concept for native English speakers. Similarly, Russian distinguishes between two families of what English speakers call \say{blue}: the lighter \foreign{goluboy} and the darker \foreign{siniy}. Reaction time experiments show the cognitive importance of these categories \citep{gilbert2006whorf, winawer2007russian},
and 
the existence of a named category both aids \citep{brown1954study} and guides \citep{bae2015some, cibelli2016sapir} color judgment and memory. 

Color terms may be \term{concrete} (i.e., derived from a real-world referent like \say{blood} or \say{sky}) or \term{abstract}. Diachronic processes can weaken the link between a concrete term and its referent, until a new cohort of speakers believes the term to be abstract. Indeed, this process explains the development of English color words \citep{casson1994englishcolor}. In addition to metonymy with named things, the words may be borrowed, compounded, or inherited from an ancestor language.

While industrialized societies' languages possess a wealth of color words \citep{hardin2014berlin}, only a handful are considered \emph{basic} color terms; the remainder are secondary. A \term{basic color term (BCT)} must satisfy four obligatory criteria \citepalias{berlin1969basic}: 
\begin{enumerate}[nosep]
    \item It must be monolexemic (and monomorphemic). \say{Light blue} and \say{blue-green} each contain two lexemes and do not qualify.
    \item It may not possess any color hypernyms (superordinate color terms). 
    (E.g., \say{lavender} has the hypernym \say{purple}.)
    \item It may not be limited in application to a narrow class of objects. \say{Blond(e)}
    may only be applied to a handful of referents like hair, wood, and beer, for example.
    \item It must be psychologically salient. This implies that the color term has
    a stable range of reference across speakers and has an entry in the lexemic inventory
    of most (if not all) native speakers' respective idiolects.
\end{enumerate}
Additional criteria are introduced in cases of doubt \cite{kay1978linguistic}, though these are subjectively applied \cite{crawford1982defining}. Among these: (5) a BCT is not the name of an object that characteristically has a particular color; in other words, the color must be \term{abstract}, and not grounded in some concrete object
(which rules out colors like gold and salmon). Additionally, (6) recent foreign loan words \say{are suspect}, and (7) if
the lexemic status of the word is difficult to judge, then multimorphemic words are also \say{suspect}. 

\begin{figure}
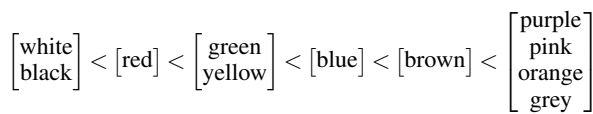

\centering
\small
\[
\begin{bmatrix}
\text{white} \\
\text{black} \\
\end{bmatrix}
<
\begin{bmatrix}
\text{red}\\
\end{bmatrix}
<
\begin{bmatrix}
\text{green}\\
\text{yellow}\\
\end{bmatrix}
<
\begin{bmatrix}
\text{blue}\\
\end{bmatrix}
<
\begin{bmatrix}
\text{brown}\\
\end{bmatrix}
<
\begin{bmatrix}
\text{purple}\\
\text{pink}\\
\text{orange}\\
\text{grey}\\
\end{bmatrix}
\]
\caption{The diachronic sequence of color acquisition \citep{berlin1969basic}.
}
\label{fig:color_order}
\end{figure}

In addition to this definition, \citepalias{berlin1969basic} surveyed speakers of 20 languages in the San Francisco Bay Area, plus a sweeping examination of the literature, to find a \term{sequence} to the emergence of color words in language. Cultures with two color words universally used them to distinguish light and warm colors from dark and cool ones; the third color was universally red, and the sequence continued until matching the set of eleven colors represented by English basic color terms. We present their partial ordering in \autoref{fig:color_order}, though later authors have proposed alterations \citep{heider1972universals, kay1975synchronic}. 

We are not the first to assess the notion of a basic color term. \citet{crawford1982defining} gives a point-by-point rebuttal on pragmatic grounds---the criteria are hard for a field worker to assess, and many introduce subjectivity that will bias data collection. \citet{lucy1997linguistics} argues that the definition provides more of a post-hoc screening tool for when the \say{denotational net} of elicitation has captured too many terms, as opposed to a morphosyntactically informed approach \citep[e.g.,][]{conklin1955hanunoo}. Finally, \citet{wierzbicka2006semantics} argues that other societies may not share the Western conception of hue-based color terms, making the application of the concept inappropriate. In addition to these postulatory objections, a vast literature of similarity judgments, reaction times, and other human measures debates the question from a cognitive perspective \citep[\emph{inter alia}]{heider1972universals, jameson2005culture, roberson2005color, roberson2008categorical,goldstein2009knowing,loreto2012origin,persaud2014influence}. 

By contrast, we examine the conditions empirically, broadly and automatically on a massively multilingual scale (versus manually and theoretically). Our evidence for assessing \citetalias{berlin1969basic}'s criteria of abstractness, monomorphemicity, and salience comes from a multilingual dragnet of color terms.

\section{Data} \label{sec:data}

We investigate the three aspects of our theory assessment---abstractness, monomorphemicity, and salience---through multilingual dictionaries.  We additionally leverage English corpora to explore abstractness and salience. We use these to construct a dataset of color senses and translations, with scores along numerous axes. As a final resource to investigate salience, we use a global elicitation of color terms from pre-industrialized societies.

In English, the basic color terms are red, orange, yellow, green, blue, purple, brown, pink, black, white, and grey. These align to the eleven basic color categories identified by \citet{berlin1969basic}. In addition to these eleven, we consider a list of 92 second-tier color terms identified by \citet{casson1994englishcolor}. These were elicited from 30 speakers over several days to ensure salience, then filtered by a dictionary to keep only conventional (rather than novel) color descriptors. We omit 12 of these which do not appear in the datasets we employ.

Translation may act as a useful resource for disambiguating word senses \citep{diab2002unsupervised}. By translating English BCTs into other languages, we can find their basic color terms. Then, back-translating to English, we obtain a list of the potential senses of a given color term.
We draw translations of color terms from two large type-level dictionary resources, PanLex \citep{baldwin2010panlex, kamholz2014panlex} and Wiktionary, which together provide color word translations for 2491 languages or dialects.

For each of the 11 basic color concepts and 80 secondary terms \(e\) in English, we translate it into every available foreign language \(\ell\) by dictionary lookup to get a set of non-English color words \(F_\ell\). We then back-translate each term into English (again by lookup) to get a set~\mbox{\(E_\ell^{(e)}=\bigcup_{f \in F_\ell} \{e' \mid \mathrm{trans}_{\ell \rightarrow \mathrm{En}}\left(f\right) = e'\}\)} of possible round-trip translations through the language \(\ell\). The final dataset contains tuples of the form (English color word, foreign language, foreign word, English back-translation).

\section{Summary of Experiments} \label{sec:experiments}

We perform a series of experiments to explore the abstractness, morphology, and salience of color terms across thousands of languages.\footnote{ We give our data and implementations at \url{https://github.com/aryamccarthy/basic-color-terms}.} In addition to qualitative observations, our experiments evaluate these qualities through several metrics, illuminating flaws in the definition of \say{basic color term}. 
When averaging the 14 features together, the implied total ordering is suggestive of the original \citetalias{berlin1969basic} sequence.

\paragraph{Goodman and Kruskal's gamma} 
We measure correlation between basicness and our features with Goodman and Kruskal's gamma \citep{goodman1954measures,goodman1959measures,goodman1963measures,goodman1972measures}, which is well suited for comparing binary variables to ordinal ones. It is a pair-counting measure which ignores tied values. We compute it by maximum likelihood estimation, giving an expression:
\begin{equation}
\gamma = \frac{N_s - N_d}{N_s + N_d}\text{,}
\end{equation}
where \(N_s\) is the number of color pairs for which basicness and a feature 
agree in their ranking; \(N_d\) is the number of pairs ranked in opposite orders.
Arbitrarily, we represent basicness as \(1\) and non-basicness as \(0\). We will be concerned only with the magnitude, not the direction of correlation. A score of \(\pm 1\) thus indicates perfect correlation with basicness.

\paragraph{A remark on noise}
Each of our measures is imperfect; it possesses a bias. 
Combining the results from many weak indicators gives robustness to the noise in any given measure. Foreshadowing future discussion, we see this supported empirically by the effect of aggregation on Goodman and Kruskal's gamma.

\section{Abstractness} \label{sec:abstractness}

\begin{table}[th!]
    \centering
    \begin{adjustbox}{width=\columnwidth}
    \begin{tabular}{l l r r l}
    \toprule
        Category & Back-translation & Freq. & Conc. & OED POS \\
    \midrule
        Black& black & 1065 & 3.76 & Adj \\
             & dark & 467 & 4.29 & Adj \\
             & dirty & 162 & 4.23 & Adj \\
    \midrule
        White& white & 4423 & 4.52 & Adj \\
             & bright & 163 & 3.92 & Adj \\
             & clear & 163 & 3.55 & Adj \\
    \midrule
        Red  & red & 3053 & 4.24 & Adj \\
             & reddish & 320 & 3.42 & Adj \(\rightarrow\) Adj \\
             & crimson & 286 & 4.00 & Adj \\
    \midrule
        Green & green & 3786 & 4.07 & Adj \\
             & unripe & 468 & 3.31 & Adj \(\rightarrow\) Adj \\
             & blue & 242 & 3.76 & Adj \\
    \midrule
        Yellow & yellow & 1306 & 4.30 & Adj \\
             & saffrony & 133 & - & N \(\rightarrow\) Adj \\
             & luteous & 94 & - & F.W. \(\rightarrow\) Adj \\
    \midrule
        Blue  & blue & 1324 & 3.76 & Adj \\
             & gloomy & 361 & 2.52 & Adj \\
             & grim & 293 & 1.82 & Adj \\
    \midrule
        Brown & brown & 671 & 4.48 & Adj  \\
             & pink & 314 & 3.93 & Adj  \\
             & brownish & 106 & 3.24 & Adj \(\rightarrow\) Adj  \\
    \midrule
        Purple & purple & 932 & 4.04 & N \\
             & violet & 324 & 4.48 & N \\
             & purpure & 148 & - & N \\
    \midrule
        Pink & pink & 121 & 3.93 & Adj \\
             & rose & 13 & 4.90 & N  \\
             & carnation & 10 & 3.93 & N  \\
    \midrule
        Orange & orange & 1257 & 4.66 & N   \\
             & orange tree & 89 & -  & N   \\
             & mandarin & 87 & 3.67 & N   \\
    \midrule
        Gray & gray & 775 & 3.46 & Adj \\
             & grey & 570 & 4.11 & Adj \\
             & greyish & 155 & 3.99 & Adj \\
        
    \bottomrule
    \end{tabular}
    \end{adjustbox}
    \caption{Top back-translations and their concreteness for each of the 11 basic color categories. Extended in the supplementary material. Part of speech is first-listed from the Oxford English Dictionary \citep{simpson1989oxford}.}
    \label{tab:concreteness}
\end{table}

The basic color terms in English did not start as abstract concepts. For instance, \say{orange} and \say{pink} were originally derived from the concrete color of a fruit (\emph{Citrus sinensis}) and flower (\emph{Dianthus plumarius}) respectively, and earlier the English color term \say{black} had its origin in a word for \emph{soot}, but the abstract color senses of these words rose in relative popularity; many supplanted the original definitions as the more common \term{word sense} \citep{casson1994englishcolor}. It could be that many \say{basic} color terms emerged metonymically from concrete, real-world referents, as in English.


\subsection{Concreteness judgments} \label{sec:word-concreteness}

As a first pass, we directly look up the concreteness for each color word on our list, in a dataset of 40,000 English lemmas' concreteness  \citep{brysbaert2014concreteness} as rated on a Likert scale by Amazon Mechanical Turk workers in the United States. 
As expected, we find that concreteness negatively correlates with basicness (\(\gamma = -0.58\)). Many non-basic colors are less concrete than basic terms; for instance, \say{beige} is the least concrete color (3.41), and most words are less concrete than \say{orange}.


\subsection{A hologeistic perspective} \label{sec:translation-concreteness}

A word may have multiple senses, which we hope to capture by taking a pan-lingual, or hologeistic, 
perspective, getting at the concept itself rather than any surface form. To do this, we find the human-judged concreteness \emph{of each of our back-translations}, then average these for each color term, weighted by the number of languages for which a word is a back translation. This balances between a single frequent sense and multiple infrequent senses. 

Performing this averaging over languages and senses magnifies our correlation to \(\gamma = -0.62\); clearly, exploiting the diversity of senses is beneficial. Still, there is no clear separation between basic and secondary color terms. Concreteness or abstractness thus provides incomplete evidence of basicness.

\subsection{Part of speech as a proxy for concreteness} \label{sec:pos}

Adjectives are, on average, perceived as more abstract than nouns \citep{darley1959scaling}. We affirm this finding: in the \citeauthor{brysbaert2014concreteness}\ judgments,  nouns are less abstract than adjectives on average (3.53 versus 2.50). 
Because of this, we are comfortable using color words' part of speech as a coarse hint of their abstractness.

We collect part of speech annotations from two sources: the Google Books Ngram Corpus, containing about 4\% of all books ever printed \citep{michel2011quantitative}, and the Penn Treebank \citep{marcus1993building}. The former is machine-annotated for part-of-speech (and thus noisier); the latter is annotated by linguists. For each corpus, we compute the ratio of adjectival relative to nominal part of speech, as well as total frequency.%
\footnote{ These features work double duty---they shed insight into both the concreteness and the salience 
as descriptive concepts of each color word. It can lead to spurious ordering, though, if not tempered by other measures---\say{gold} and \say{flesh} are both frequent because of their material senses, so raw frequency is not an indicator of basicness.}

\section{Morphology} \label{sec:morphology}

In this section, we ask whether there are affixes that are highly correlated with color; these can be either general derivational affixes or sequences specific to color terms, as in \autoref{tab:intro} and \autoref{tab:recipes}. 
The presence of subword structure in basic concepts' translations would show that they violate the monomorphemicity criterion.

Although the \citetalias{berlin1969basic} criteria demand monolexemic and monomorphemic words, color terms in many languages are formed by some derivational process from a concrete term. To discover the components in an unsupervised fashion, as is necessary for languages in the long tail of linguistic resource availability, we look for constituent morphemes and constituent lexemes by segmentation and compound detection, respectively, on each foreign language of our dataset, then use these to score colors' basicness.  

\subsection{Affix discovery} \label{sec:affix-discovery}

Segmentation and affix discovery is a challenge for the low-resource languages in our study. To give signal to the model, we leverage our other metrics. We compute  the percentage of the time that a color word's translation occurs with a suffix that is strongly associated with one of the top 10-highest-ranking colors on the basicness scale, according to the aggregation we mention in \autoref{sec:experiments} and detail in \autoref{sec:aggregation}. In other words, words that associate with a typical color affix in that language tend to be colors.

This is not a test for basicness, per se, but rather being a likely color, so it supplements the part of speech measures. But it does diminish the rank of almost all words with senses/translations that are not primarily colors. In addition, this measure has the highest correlation of any of ours with basicness (\(\gamma = 0.92\))---though this is not surprising, as it was computed by bootstrapping from the other results, which already correlate well in aggregate.

As another tack, we identify likely color-related and general derivational affixes by \term{unsupervised morphological segmentation}. 
We define a probability distribution over segmentations. Let \(S = s_1 s_2 \ldots s_m\) segment the word \(W = {\textsc{bos}} w_1 w_2 \ldots w_n {\textsc{eos}}\). We seek to find the optimal \(S^*\). To do this, we decode a model
\[S^* = \arg\max_{S} \Pr(S) = \arg\max_{S} \prod_{s_i \in S} \Pr(s_i)\]
using the Viterbi algorithm, where the individual segment probabilities are maximum a posteriori estimates under a Dirichlet prior (\(\alpha = 0.01\)). The model, inspired by \citet{ge1999discovering}, is similar to other unigram segmentation models \citep{creutz2005unsupervised, kudo2018subword}. We then search for these affixes across the terms recorded in that language, to determine whether the affix is broadly derivational or specific to color terms. Select results are given in \autoref{tab:affixes}.

Consider the \foreign{-tic} suffix in Nahuatl (\lang{nci}). This morpheme  semantically denotes \say{color}; one combines real-world referents with this suffix to obtain colors, e.g., \foreign{chichiltic} (chili-color). This appears even with black and white---the first basic color terms in \citetalias{berlin1969basic}'s ordering---implying that Nahuatl lacks basic color terms under the \citetalias{berlin1969basic} definition.

We also see derivational morphemes, which are applied to words from a given part-of-speech class to convert them to another class---e.g., \say{\foreignlanguage{russian}{\emph{тут}}} in Archi (\lang{aqc}) in Table \ref{tab:affixes}, which is a fused morpheme denoting adjectivalization and marking Archi's fourth gender. This morpheme appears in the Archi's terms for black and white. As with Nahuatl, this implies that Archi lacks basic color terms---according to a strict interpretation of the criteria.

\begin{table}[t]
    \centering
    \begin{tabular}{l l}
    \toprule
        Language & Affixes \\
    \midrule
        \lang{nci} & -tic\(\dagger\) (40\%), tla- (27\%), xox- (27\%) \\
        \lang{aqc} & -\foreignlanguage{russian}{тут}\(^*\) (36\%)\\
        \lang{dbu} & -\textipa{N}g\'o (29\%)\\
        \lang{mpj} & -lpa\(^*\) (26\%), -ku (15\%)\\
        \lang{ciw} & -aan-\(\dagger\) (23\%), -zo\(^*\) (23\%) \\
    \bottomrule
    \end{tabular}
    \caption{Common example affixes. \ensuremath{*} indicates a general derivational affix. \ensuremath{\dagger} indicates a color-specific morpheme. Bare affixes are neither. Included for each morpheme is the percentage of color words in our dataset in the given language which contain the morpheme. ISO language codes are given in \autoref{sec:lang-codes}.}
    \label{tab:affixes}
\end{table}

\subsection{Compound detection} \label{sec:compound-detection}
\begin{table}[t]
    \centering
    \resizebox{\columnwidth}{!}{
    \begin{tabular}{l l}
    \toprule
        Method & Example \\
    \midrule
        concrete + color & \lang{cmn}: \begin{CJK}{UTF8}{gbsn}橙\end{CJK} (orange) + \begin{CJK}{UTF8}{gbsn}色\end{CJK} (color) \\
        color + der.\ affix & \lang{spa}: anaranja (orange) + do \\
        Adj.\ + color & \lang{deu}: dunkel (dark) + rot (red) \\
    \bottomrule
    \end{tabular}
    }
    \caption{Discovered concatenation strategies pooled across languages, with representative examples from Mandarin, Spanish, and German.}
    \label{tab:recipes}
\end{table}

To particularly identify color names which are formed by compounding of words, we extend a model for compound discovery to identify color terms which were produced compositionally. This lets us ask two questions about the BCT definition: (1) Across languages, are there \say{basic} color terms that are not monolexemic, and (2) Are \say{basic} terms less likely to be compounds? The answer to both, we find, is \say{yes}.

\citet{wu2018massively} propose a multilingual compound analysis and generation method that only requires a readily available multilingual dictionary. They first extract potential compounds by splitting any word into three substrings corresponding to a left component, glue, and right component. These compounds are used to construct compound ``recipes''. For example, they discovered that the concept of `hospital' is frequently represented across a variety of languages as a compound of `sick'/`disease' and `house'/`home' in their respective language. They use these recipes to score their initial list of potential compounds, filter out low-scoring, unlikely compounds, and performing a second pass of recipe construction, resulting in a higher-quality compound dataset. Compound analysis is performed in a similar manner as recipe construction. Compound generation takes into account language-specific knowledge of which components and glues are common in that language.

\citet{wu2018massively} consider only single- or zero-character  glue between the two components. By contrast, \term{we allow glues of arbitrary length}, exhaustively searching through all segmentations into three parts. This increases the algorithmic complexity by \(\Theta(K)\), where \(K\) is the length of the word. Searching through our PanLex and Wiktionary foreign translations of only our basic color terms, we find several examples of compounded words. Some examples are given in \autoref{tab:recipes}. The frequency of a color being expressed by compounding, though, turns out to be a weak indicator of basicness (\(\gamma = 0.35\)).


\subsection{An aside on borrowings} \label{sec:borrowings}
\begin{table}
    \centering
    \begin{tabular}{l r r}
    \toprule
        Process & Basic & Secondary \\
    \midrule
        Inheritance & 1161 & 2356 \\
        Derivation & 82 & 183 \\
        Cognate & 303 & 483 \\
        Borrowing & 18 & 84 \\
        None of these & 42566 & 65969 \\
    \bottomrule
    \end{tabular}
    \caption{Sources of foreign color words. The colors are not mutually exclusive. 
    }
    \label{tab:sources}
\end{table}

One of \citetalias{berlin1969basic}'s lower-tier criteria for color terms was that borrowings were \say{suspect}. Here we examine how often color terms are borrowed
, as well as other avenues for color construction.

In addition to translations and definitions, Wiktionary provides etymologies for many languages. 
These relations have been parsed and extracted as EtymDB \citep{sagot2017extracting}.

We report the aggregation of EtymDB's parsed etymologies in \autoref{tab:sources}; these are broken down by color in the supplementary material. 
Basic colors are more often created by borrowing, suffixing, and compounding than secondary colors; nevertheless, this should be taken with a grain of salt: The annotations for secondary colors are less complete, so while we use these scores, we do not take the criterion of borrowings as a prong of our criticism. 

\section{Salience} \label{sec:salience}

\citetalias{berlin1969basic} assessed salience by a color's tendency to appear earlier when asking speakers to list their language's colors. More general assessments are outlined by \citet{hays1972color}: word length, frequency of use, ethnographic frequency, and correlation of vocabulary size with cultural complexity. While the final one of these is beyond our scope, we present simple experiments to test the other three, based on our translation dataset. 

\subsection{Word length} \label{sec:word-length}
\citet{durbin1972} supposed, based on Zipf's Law of Abbreviation \citep{zipf1932selected, zipf1949human}, that word length would decrease for more salient, broadly used color terms. We test this across over 2400 languages by computing the mean length of all translations for each English color word, regardless of script. 

With the exception of grey, the first six colors of the \citepalias{berlin1969basic} sequence---black white, red, yellow, green, and blue---have lower average lengths than the subsequent five. This supports \citeauthor{durbin1972}'s two-phase theory of basic color terms. Still, beyond this handful, there is no clear separation between basic and non-basic colors. There is only a moderate correlation (\(\gamma = -0.41\)) with basicness.

\subsection{Frequency: Usage and ethnography}

 \begin{figure}[t]
     \centering
     \includegraphics[width=\linewidth]{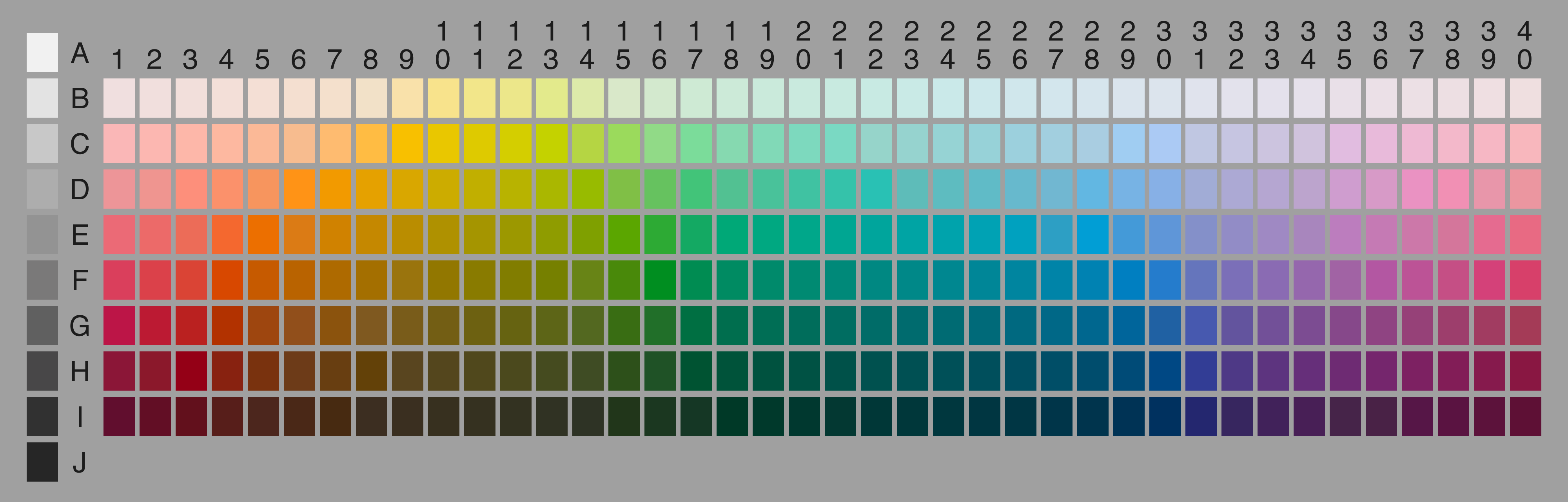}
     \vspace{0.5em}
     \includegraphics[width=\linewidth]{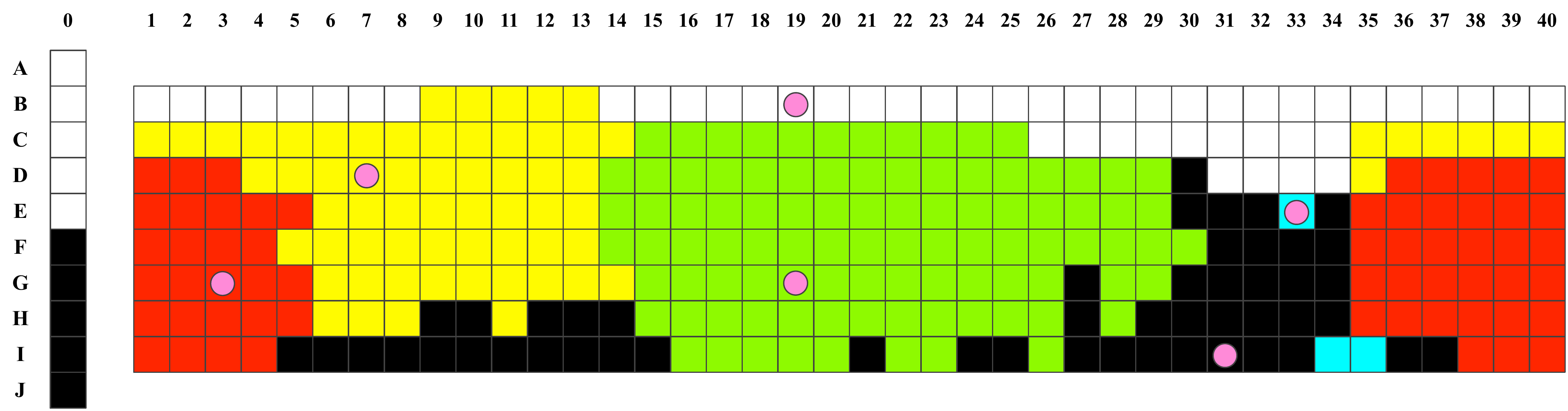}
     \caption{\emph{Top:} The stimulus palette used to collect color namings by \citet{berlin1969basic} and later the World Color Survey. Colors vary vertically in lightness and horizontally in hue. All are fully saturated. \emph{Bottom:} The extensions of the six colors identified by one speaker of the Iduna language, spoken in Papua New Guinea.}
     \label{fig:stimulus_palette}
 \end{figure}

\begin{figure}[t]
    \centering
    \includegraphics[width=\linewidth]{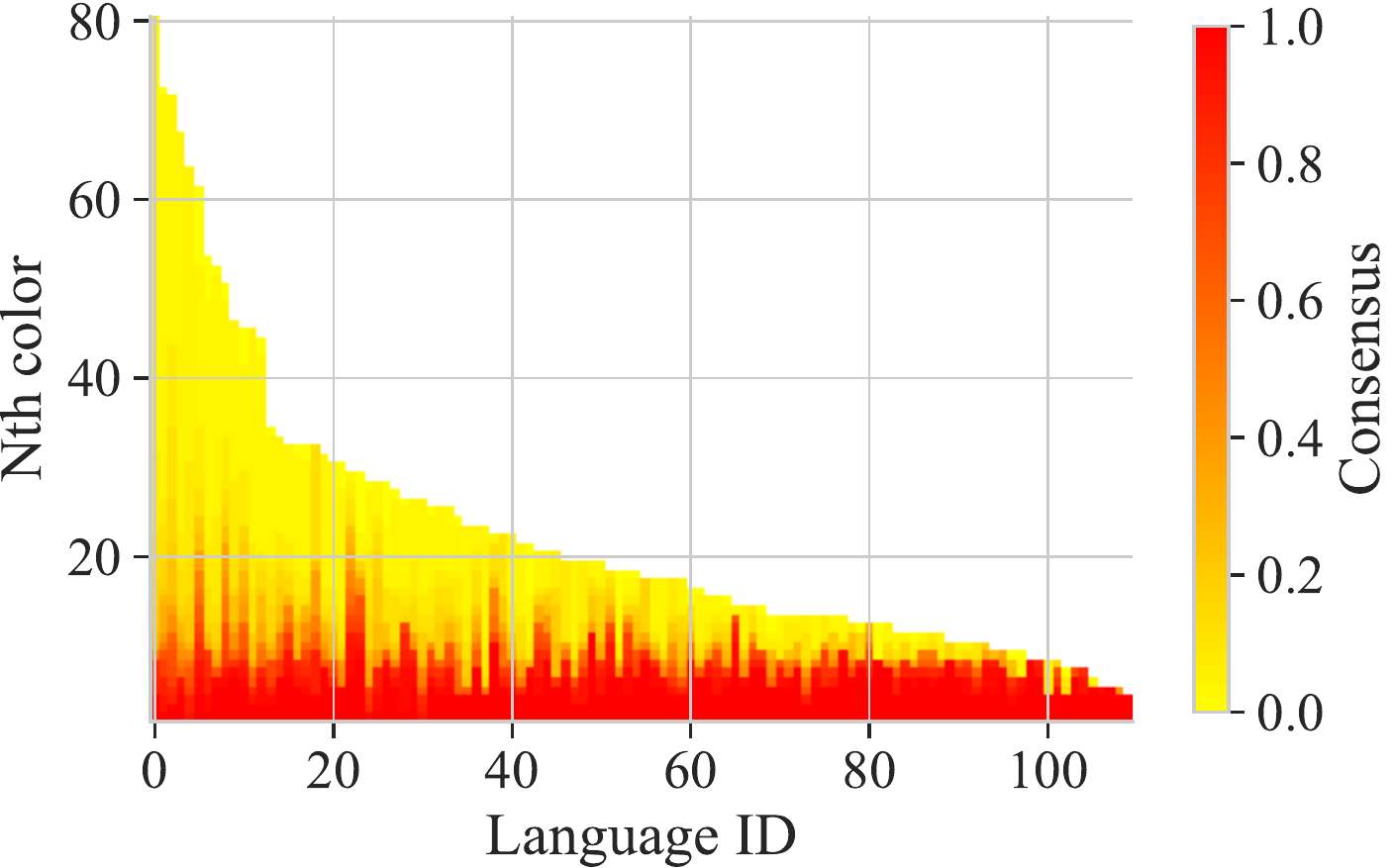}
    \caption{The relative frequencies of each language's color term in the World Color Survey. The height of the colored columns forms a histogram, showing the total number of elicited words. Redder color indicates broader consensus: The word was elicited by a greater fraction of the language's speakers.}
    \label{fig:heterogeneity}
\end{figure}


Neither English corpora nor multilingual dictionaries give a complete picture of the world's languages; after all, only 3 to 4 thousand of them have writing systems \citep{lewis2015ethnologue}.
%
To augment our analyses, we consider the grounded speaker elicitation performed in the World Color Survey \citep[WCS;][]{cook2005world}. In it, 2568~speakers from 110 pre-industrialized societies gave their color naming judgments about colored chips on a stimulus palette (see \autoref{fig:stimulus_palette}) 
which evenly varied hue and brightness. Field workers then transcribed the utterances and applied the \citetalias{berlin1969basic} criteria to ascertain which colors were basic.
%
With the WCS, we can discover the synchronic homogeneity of a term's use for the labeling task.

\paragraph{Do all speakers agree that a given term should be used?} We examine the consensus of use among words elicited in the World Color Survey. For each elicited color word in the language, we count the number of speakers who used it. 
Indeed, the distribution is varied. In \autoref{fig:heterogeneity}, we give a histogram of the total number of color terms elicited from each language, as well as the consensus for each individual color. We see a lexical core and periphery for most languages. Most color terms are unique, and languages may have up to 79 terms given among their speakers. It is natural for a speaker to use an unexpected word, still believing the color to fit in a more basic category (as with `turquoise' and `blue' in English), but it is surprising that these words would not be sifted out by field linguists' \say{denotational net}.

\paragraph{Do all speakers have the same number of colors?} We look at the variability of sizes of each speaker's inventory for a language, rather than the consensus on each term. The standard deviation of inventory sizes from the speakers of each language shows notable variation, especially when the mean inventory size exceeds 6.

Heterogeneity in within-language color inventories is unsurprising; as \citet{kay1975synchronic} noted, younger members of numerous language communities possess a broader inventory of basic color terms, following \citeauthor{berlin1969basic}'s evolutionary sequence of inventories. Nevertheless, basic color terms are defined on a per-language level; they ignore this synchronic variation. This raises questions for future work: Are there categories salient only to young speakers, not the old? What would be a population's (or language's) basic color terms?

\section{Aggregation of Features} \label{sec:aggregation}

\begin{table}
\centering
\begin{adjustbox}{max width=\linewidth}

\begin{tabular}{llrr}
\toprule
Score & Defined &     Basic &  Sequence \\
\midrule
Word concreteness &\autoref{sec:word-concreteness}  &  0.578 &  0.554 \\
Translation concreteness &\autoref{sec:translation-concreteness}        &  0.616 &  0.583 \\
Ngram frequency &\autoref{sec:pos}     &  0.897 &  0.896 \\
Ngram \%ADJ &\autoref{sec:pos}      &  0.806 &  0.780 \\
Penn TB \%ADJ &\autoref{sec:pos}           &  0.707 &  0.705 \\
Count of compounding &\autoref{sec:compound-detection} &  0.874 &  0.858 \\
Frequency of compounding &\autoref{sec:compound-detection}  &  0.354 &  0.383 \\
Affix presence &\autoref{sec:affix-discovery}        &  0.924 &  0.892 \\
Borrowing etym.\ &\autoref{sec:borrowings}          & \(-\)0.254 & \(-\)0.231 \\
Cognate etym.\ &\autoref{sec:borrowings}           &  0.797 &  0.804 \\
Derivation etym.\ &\autoref{sec:borrowings}            &  0.830 &  0.805 \\
\quad - Suffix derivation\ &\autoref{sec:borrowings}        &  0.877 &  0.869 \\
Inheritance etym.\ &\autoref{sec:borrowings}            &  0.765 &  0.774 \\
Word length &\autoref{sec:word-length}           &  0.414 &  0.427 \\
\midrule
Aggregate &\autoref{sec:aggregation}             &  0.964 &  0.962 \\
\bottomrule
\end{tabular}

\end{adjustbox}
\caption{Goodman and Kruskal's gamma rank correlation between each measure and both basicness and the diachronic color sequence. Features not thought to indicate basicness (as discussed) have been negated. 
As ordering the sequence is more nuanced than sifting basic from non-basic, one might expect consistently lower correlations. However, the sequence gamma can have larger magnitude if, despite poor sifting, the order of the basic terms is correct.}
\label{tab:gamma}
\end{table}

We have operationalized some \citetalias{berlin1969basic} criteria for basic color terms. Independently, each fails to match up with the known set of basic color terms. Now, we aggregate our independent scores to create a robust measure of basicness from the weak measures.

We do not cherry-pick our measures; some that we include actually harm the ordering. Instead, we operationalize each criterion, then see what shakes out. We take an unweighted average of normalized scores for the same reason.
This makes the result more evocative; they come purely from our operationalization, rather than targeted 
tuning.\footnote{ Nonetheless, we characterize the contribution of individual features  in \autoref{sec:diagnostics}.}


Using the aggregated measure produces the highest correlation with both basicness and the order of the \citetalias{berlin1969basic} sequence; see \autoref{tab:gamma}. We recover the first six colors in the evolutionary sequence from \autoref{fig:color_order}: white and black, red, green and yellow, and blue! 
An extended ordering is given in \autoref{tab:total-ordering}; it suggests that the most primary of the non-basic terms are gold, scarlet, crimson, and beige.

Some further inquiry is possible. We see that orange is the 24th ranked color out of 91. This is a stark separation from the ten other basic color terms; beyond this, it has the highest concreteness of the basic terms. 

\begin{table}[t]
\centering

\begin{tabular}{lllr}
\toprule
Color & Rank  & \citetalias{berlin1969basic} &  Agg. score  \\
\midrule
\textbf{white}      & 1 & 1--2  & 1.00 \\
\textbf{black}      & 2 & 1--2 & 0.97 \\
\textbf{red}        & 3 & 3 & 0.92 \\
\textbf{green}      & 4 & 4--5 & 0.83 \\
\textbf{yellow}     & 5 & 4--5 & 0.80 \\
\textbf{blue}       & 6 & 6 & 0.79 \\
\textbf{gray}       & 7 & 8--11 & 0.73 \\
gold                & 8 &  & 0.67 \\
\textbf{brown}      & 9 & 7 & 0.66 \\
 \textbf{pink}       &10 & 8--11 &  0.64 \\
 scarlet            & 11 & & 0.64 \\
 \textbf{purple}     &12 & 8--11 &  0.62 \\
 crimson            & 13 & & 0.60 \\
 beige              & 14 & & 0.58 \\
 silver             & 15 & & 0.55 \\
 blond              & 16 & & 0.54 \\
 tan                & 17 & & 0.52 \\
 amber              & 18 & & 0.52 \\
 flesh              & 19 & & 0.51 \\
 bronze             & 20 & & 0.48 \\
\bottomrule
\end{tabular}

\caption{Total ordering of colors according to our aggregate score, with conventionally basic colors bolded. Complete results are given in the supplementary material. This includes the position of the missing basic color, orange.}
\label{tab:total-ordering}
\end{table}

\section{Discussion}  
%
%

While the notion of a basic color term has been widely used, its validity has been taken as given. With NLP techniques in the broadest multilingual survey by far, we add to the literature investigating the definition of basic color terms. The ability to produce color templates shows that monomorphemicity is an unreasonable criterion. The concreteness of many color words' back-translations violates abstractness. Finally, the heterogeneity of color naming data contends with the salience requirement. None of the traditional criteria for basic color terms hold up robustly. Despite this, when taken in aggregate (and operationalized as we do), they suggest the traditional sequence of color terms and a coarse division between basic and non-basic colors. 

As color terms are often decomposable, we can turn the decomposition on its head to \emph{generate} missing color words. We have shown cross-lingual patterns of word formation that future work can exploit, giving plausible entries in a bilingual dictionary \citep{wu2018massively}: Without the word for \say{hospital}, one can convey the concept by \say{sick}+\say{house}; likewise, without the word for \say{gray}, one can use \say{ash}+\textsc{derivational affix}. Future work will investigate generation and validation of unseen color terms. 


Finally, given that the divide between basic and secondary color terms is so blurred, future computational models of these should employ models of graded membership, such as fuzzy set theory.

\section{Conclusion}

 This paper has investigated the universal basic color term theories of \citet{berlin1969basic} and others. It provides empirically-grounded computational linguistic metrics with evidence from 2491 languages, harnessing multiple on-line resources of varying quality. We have shown that although the obligatory criteria do not in fact cleanly separate basic from non-basic colors, our features' aggregation correlates strongly with the \citeauthor{berlin1969basic} basic/secondary color term partition (\(\gamma=0.96\)). The aggregation also largely predicts the \citeauthor{berlin1969basic} hypothesized universal acquisition sequence, which is in no way directly entailed by the basicness criteria. Thus, we provide further empirical evidence from computational linguistics in support of the \citetalias{berlin1969basic} claims, while also providing additional nuance and perspective thereon.
 
 

\section*{Acknowledgments}
We thank several members of the Johns Hopkins University Center for Language and Speech Processing, as well as Elyse Wilson, for early discussions that shaped the manuscript. Keehoon Trevor Lee prepared part of the data for \autoref{tab:concreteness} as an undergraduate research assistant. We thank Ryan Cotterell for producing the lower part of \autoref{fig:stimulus_palette} and Dorothy Hu for presentation suggestions on \autoref{fig:heterogeneity}. 

\bibliography{naaclhlt2019}
\bibliographystyle{acl_natbib}

\clearpage

\appendix

\section{Language codes} \label{sec:lang-codes}

We use an internationally standard set of codes (ISO 639-3) to represent languages. Those for languages referenced in the main text are:

\begin{center}
\begin{tabular}{l r}
    \toprule
    Language & ISO 639-3 \\
    \midrule
    Archi & \lang{aqc} \\
    Bondum Dogon & \lang{dbu} \\
    Cantonese & \lang{yue} \\
    Chippewa & \lang{ciw} \\
    German & \lang{deu} \\
    Italian & \lang{ita} \\
    Korean & \lang{kor} \\
    Mandarin Chinese & \lang{cmn} \\
    Martu Wangka & \lang{mpj} \\
    Nahuatl & \lang{nci} \\
    Russian & \lang{rus} \\
    Spanish & \lang{spa} \\
    Welsh & \lang{cym} \\
    \bottomrule
\end{tabular}
\end{center}

\section{Feature importances} \label{sec:diagnostics}

While it is encouraging that the diachronic acquisition sequence of black and white, red, and so on emerges from our operationalization of the basicness criteria, we may be interested in some diagnostic information about which features are not actually informative, as well as clues to the robustness of our findings.

To do this, we perform recursive feature elimination \citep{guyon2002gene}. The resultant feature sets for correlating with basicness and acquisition sequence, respectively, are:
\begin{description}
    \item[Basicness] 9 features: Word concreteness, Count of compounding, Frequency of compounding, Cognate etymology, Derivation etymology, Google Ngram frequency, Google Ngram percentage adjectival, Penn Treebank percentage adjectival, and Affix presence. Gamma is 0.983, which closes the gap to perfect correlation by about 50\%.
    \item[Acquisition sequence] 6 features: Count of compounding,
Frequency of compounding, Suffix derivation, Google Ngram frequency, Penn Treebank percentage adjectival, and Affix presence. Gamma is 0.988, which improves the gap to perfect correlation by about 70\%.
\end{description}
In the latter case, the greatest benefit is derived from improving the rank of \emph{orange}. Namely, it moves from position 24 to position 15 in the ranking, while the first six colors hold their positions. \emph{Brown}'s position also improves, swapping with \emph{gray}.

Interpreting this in light of our operationalization, we may ask which of the criteria are most pertinent. We keep three morphological features, one abstractness feature, and two hybrid features (the frequencies) which also convey salience and are tempered by the other features (so that, say, \emph{gold} and \emph{flesh} aren't inappropriately ranked---see \autoref{sec:pos}). Thus, it seems that we cannot eliminate any of the three categories of features we have created, without harming our correlation.

\end{document}